\documentclass[preprint,12pt]{elsarticle}

\usepackage{amssymb}
\usepackage{amsmath}

\usepackage{lineno}

\usepackage[pagebackref,breaklinks,colorlinks]{hyperref}
\usepackage{cleveref}
\usepackage{booktabs, caption, tabularx}

\usepackage{listings}
\usepackage{tcolorbox}
\newtcolorbox{promptbox}[1]{
    colback=gray!10,             
    colframe=blue!50,            
    fonttitle=\bfseries\scriptsize,
    title=#1,
    fontupper=\tiny,
    boxrule=0.5pt,               
    arc=3pt,                     
    left=5pt, right=5pt,         
    top=3pt, bottom=3pt,         
    boxsep=5pt                   
}
\usepackage{bbding}

\journal{Image and Vision Computing}

\begin{document}

\begin{frontmatter}



\title{MITS: A Large-Scale Multimodal Benchmark Dataset for Intelligent Traffic Surveillance}


\author[a1,a2]{Kaikai Zhao}
\author[a1,a2]{Zhaoxiang Liu\corref{*}}
\ead{liuzx178@chinaunicom.cn}
\author[a1,a2]{Peng Wang}
\author[a1,a2]{Xin Wang}
\author[a1,a2]{Zhicheng Ma}
\author[a1,a2]{Yajun Xu}
\author[a1,a2]{Wenjing Zhang}
\author[a1,a2]{Yibing Nan}
\author[a1,a2]{Kai Wang}
\author[a1,a2]{Shiguo Lian\corref{*}}
\ead{liansg@chinaunicom.cn}

\cortext[*]{Corresponding author(s)}

\affiliation[a1]{organization={Unicom Data Intelligence, China Unicom},
            city={Beijing},
            postcode={100033}, 
            country={PR China}}

\affiliation[a2]{organization={Data Science \& Artificial Intelligence Research Institute, China Unicom},
            city={Beijing},
            postcode={100033}, 
            country={PR China}}


\begin{abstract}
General-domain large multimodal models (LMMs) have achieved significant advances in various image-text tasks. However, their performance in the Intelligent Traffic Surveillance (ITS) domain remains limited due to the absence of dedicated multimodal datasets. To address this gap, we introduce \textbf{MITS} (Multimodal Intelligent Traffic Surveillance), the first large-scale multimodal benchmark dataset specifically designed for ITS. MITS includes \textbf{170,400 independently collected real-world ITS images} sourced from traffic surveillance cameras, annotated with \textbf{eight main categories} and \textbf{24 subcategories} of ITS-specific objects and events under diverse environmental conditions. Additionally, through a systematic data generation pipeline, we generate \textbf{high-quality image captions} and \textbf{5 million instruction-following visual question-answer pairs}, addressing \textbf{five critical ITS tasks}: object and event recognition, object counting, object localization, background analysis, and event reasoning. 
To demonstrate MITS’s effectiveness, we fine-tune mainstream LMMs on this dataset, enabling the development of ITS-specific applications. Experimental results show that MITS significantly improves LMM performance in ITS applications, increasing LLaVA-1.5's performance from 0.494 to 0.905 (+83.2\%), LLaVA-1.6's from 0.678 to 0.921 (+35.8\%), Qwen2-VL's from 0.584 to 0.926 (+58.6\%), and Qwen2.5-VL's from 0.732 to 0.930 (+27.0\%). We release the dataset, code, and models as \href{https://github.com/UnicomAI/UnicomBenchmark/tree/main/Multimodal-Intelligent-Traffic-Surveillance}{open-source}, providing high-value resources to advance both ITS and LMM research.
\end{abstract}




\end{frontmatter}


\section{Introduction}
\label{sec:intro}

Intelligent Traffic Surveillance (ITS) systems ~\cite{8771378} enhance traffic efficiency and safety by monitoring, analyzing, and managing real-world traffic conditions. Artificial intelligence-driven visual algorithms play a critical role in ITS by processing surveillance images to enable automated traffic analysis and decision-making.
Traditionally, most ITS applications rely on convolutional and recurrent network-based ``small” models for tasks such as image classification, object detection, tracking, and segmentation~\cite{9190063,9198908,tang2017vehicle,xu2022tad,xu2024raod,XIA2024105336,PAN2024105276,YANG2018143,LEE201924}.

Despite their success in specific tasks under constrained conditions, these small models exhibit several critical limitations. First, their computational constraints hinder robustness in complex traffic environments ~\cite{faghri2023reinforce,WANG2025105428}. Second, their recognition capability is restricted to a limited set of predefined categories, severely restricting their scalability. Third, they often require retraining for new tasks or scenarios due to limited generalization. Finally, as unimodal models, they lack efficient multimodal interaction capabilities, further constraining their applicability. These limitations significantly impede the advancement of higher-level intelligence in real-world ITS systems.

\begin{figure*}[htbp]
  \centering
   \includegraphics[width=1.0\textwidth]{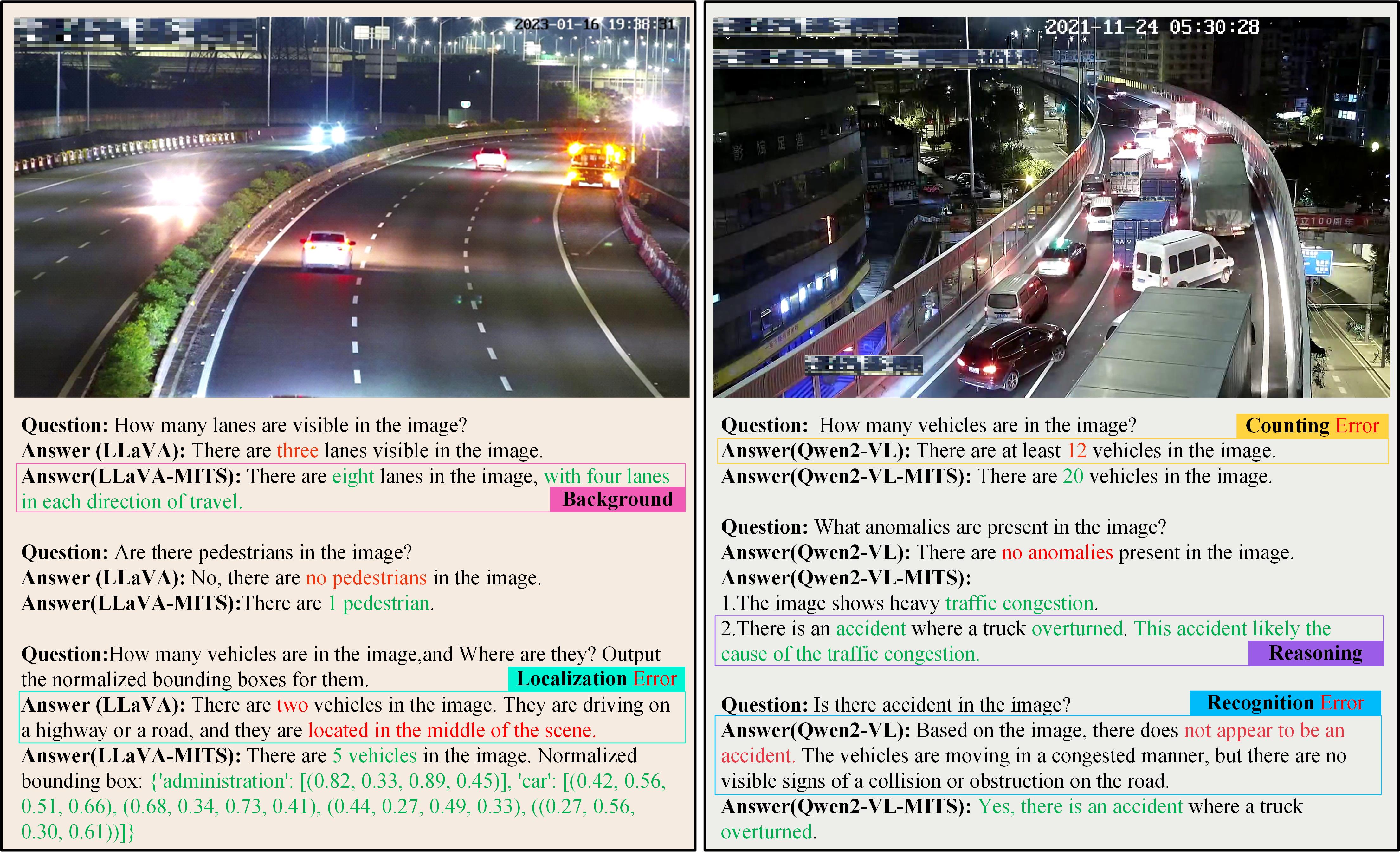}
   \caption{Performance comparison between original and fine-tuned models in ITS. LLaVA and Qwen-VL series serve as original general-domain models and are fine-tuned on our MITS dataset. Red text indicates incorrect answers from the original models, including recognition, counting, and localization errors, while green text highlights the corrections made by the fine-tuned models. MITS also enhances the models' capabilities in background analysis and reasoning within ITS.}
   \label{fig:badcase}
\end{figure*}

In contrast, large multimodal models (LMMs), particularly large vision-language models ~\cite{zhu2023minigpt,liu2024visual,Qwen-VL,Qwen2VL,li2024llava,bai2024m3d,marcu2023lingoqa,cao2024maplm,ZHA2025105484,zhu2025continual,sun2024vrp,sun2022singular,tang2025visual,tang2023context,sun2025exploring,LI2025105571,PAULRAJ2024105139,HONG2025105628,ZHANG2025105610,LI2024105172}, offer superior computational capacity, enhanced recognition and understanding capabilities, flexible deployment and interaction mechanisms, strong zero-shot generalization, and efficient scalability. 
While general-domain LMMs ~\cite{zhu2023minigpt,liu2024visual,Qwen-VL,Qwen2VL} demonstrate remarkable adaptability, specialized applications in vertical domains such as medical imaging ~\cite{li2024llava,bai2024m3d,WANG2024105069,GUARRASI2025105509,JIANG2025105463,PULARI2025105649} and autonomous driving ~\cite{marcu2023lingoqa,cao2024maplm,ZHA2025105484} often require domain-specific models. Similarly, applying general LMMs directly to ITS tasks often results in suboptimal performance. As illustrated in \Cref{fig:badcase}, general models like LLaVA and Qwen2-VL frequently produce recognition, counting, and localization errors, highlighting the limitations of general-domain LMMs and underscoring the necessity of domain adaptation. These issues stem from unique scene variations and inherent semantic alignment challenges in ITS ~\cite{li2024llava,cao2024maplm}, which hinder the effectiveness of general models.

To our knowledge, no ITS-specific LMMs have been developed, primarily due to the absence of high quality, domain-specific instruction-following data. To bridge this gap, we introduce MITS (Multimodal Intelligent Traffic Surveillance), the first large-scale multimodal benchmark dataset specifically designed for ITS. MITS comprises 170,400 independently collected real-world ITS images with high quality captions and 5 million instruction-following visual question-answer (VQA) pairs, enabling comprehensive evaluation and advancing multimodal learning in ITS applications.

Our primary contributions are summarized as follows:

\textbf{1. Large-scale ITS multimodal dataset.} We introduce MITS, the first large-scale multimodal benchmark dataset tailored for ITS. Specifically, (1) We collect 170,400 images from real-world ITS cameras, annotated for classification and detection across eight categories and 24 subcategories of objects and events. Given the inherent challenges in acquiring road surveillance camera data, the collection and curation of 170,400 ITS images themselves constitutes a significant contribution to ITS research and applications. (2) We develop a multimodal data generation pipeline to produce high-quality image captions and 5 million instruction-following VQA pairs, addressing five critical ITS tasks: object and event recognition, object counting, object localization, background analysis, and event reasoning. The MITS dataset provides a foundational resource for future multimodal learning research in the ITS domain.

\textbf{2. ITS-specific model adaptation and evaluation.} We fine-tune state-of-the-art multimodal models on MITS, adapting them for ITS-specific applications. Extensive experiments demonstrate the effectiveness of MITS, yielding significant performance improvements of 83.2\% for LLaVA-1.5-7B, 35.8\% for LLaVA-1.6-7B, 58.6\% for Qwen2-VL-7B, 27.0\% for Qwen2.5-VL-7B, and 56.4\% for Qwen2.5-VL-3B.

\textbf{3. Open-source.} We publicly release the dataset, code, and fine-tuned LMMs to facilitate further research and development in ITS-specific multimodal learning and evaluation. This dataset constitutes a high-value resource that simultaneously benefits both the ITS and LMM communities.

\section{Related Work}
\label{sec:relatedwork}
To the best of our knowledge, no LMMs or multimodal datasets have been specifically developed for the ITS domain. 
Therefore, this section first reviews deep learning-based small models and datasets used in ITS. 
Next, we discuss the key differences between ITS and autonomous driving (AD), followed by an overview of AD-related models and datasets. 
Finally, we analyze LMMs and multimodal datasets in other specialized domains to highlight the gap in ITS-specific multimodal learning. 
\Cref{tab:relatedwork} illustrates the key differences between our MITS and most related datasets.

\begin{table}[!ht]
    \centering
    \caption{Comparison of related traffic datasets. \textbf{Perspective:} the view of image/video, including Third-Person (TP), Ego-View (EV), Handheld (HH), and Multi-View Correlation (MVC). \textbf{\#Clip:} number of video clips. \textbf{\#Frame:} number of images. \textbf{\#TT:} number of task type. \textbf{\#OT:} number of focused objects. \textbf{Modal:} data modal, such as Vision-only (V) and Vision-Language (VL). \textbf{AT:} annotation type, such as image/video classification (CLS), object detection (OD), image segmentation (Seg), caption and visual question-answer (VQA). \textbf{AM:} annotation method, such as manual (M), and AI+Manual (A+M). }
    \resizebox{\textwidth}{!}{
    \begin{tabular}{cccccccccccc}
    \hline
        \textbf{Dataset} & \textbf{Domain} & \textbf{Source} & \textbf{Perspective} & \textbf{\#Clip} & \textbf{\#Frame} & \textbf{\#TT} & \textbf{\#OT} & \textbf{Modal} & \textbf{\#Annotation} & \textbf{AT} & \textbf{AM} \\ \hline
        TAD\cite{xu2022tad} & ITS & suverillance & TP & 400 & 360K & 1 & 4 & V & 360K & CLS,OD & M \\ 
        RAOD\cite{xu2024raod} & ITS & suverillance & TP & 557 & 19K & 1 & 10 & V & 19K & Seg & M \\ 
        LingoQA\cite{marcu2023lingoqa} & AD & public & EV & 28K & 140K & 4 & 8 & VL & 419K & VQA & A+M \\ 
        SUTD\cite{xu2021sutd} & AD & internet & HH,EV & 10K & / & 6 & 7 & VL & 62K & VQA & M \\ 
        WTS\cite{wts} & AD & public & MVC & 6K & 52K & 4 & 7 & VL & 6K & Caption & A+M \\ 
        RoadSocial\cite{Parikh_2025_CVPR} & AD & internet & HH,EV,TP & 13K & 14M & 11 & / & VL & 260K & VQA & A+M \\ 
        MM-AU\cite{Fang_2024_CVPR} & AD & internet & EV & 11K & / & 1 & 6 & VL & 58K & VQA & A+M \\ 
        MITS & ITS & suverillance & TP & 8K & 170K & 5 & 24 & VL & 5M & VQA & A+M \\ \hline
    \end{tabular}
    }
    \label{tab:relatedwork}
\end{table}

\subsection{Small Models and Datasets for ITS}
In the ITS domain, most existing studies utilize deep learning-based small models to handle tasks such as image classification, object detection, image segmentation, object tracking, and anomaly detection for traffic surveillance. Some studies ~\cite{shah2018cadp,fang2019dada,xu2022tad,XIA2024105336} propose datasets for traffic accident analysis and employ convolutional neural networks for accident detection and prediction. Other works ~\cite{zhu2016traffic,mogelmose2015detection} focus on traffic sign detection and recognition to improve traffic efficiency. The UA-DETRAC dataset~\cite{wen2020ua} offers a benchmark for multi-object detection and tracking in traffic surveillance. The RAOD dataset~\cite{xu2024raod} addresses road abandoned object detection in video surveillance, advancing object detection and segmentation tasks. Lastly, Wan et al.~\cite{9190063} propose a long video event retrieval method for efficient and accurate abnormal event retrieval in ITS.

\subsection{Models and Datasets for Autonomous Driving}
\subsubsection{Differences between AD and ITS}
Autonomous driving (AD) and intelligent transportation systems (ITS) are both critical domains within transportation research but differ significantly in data sources, camera perspectives, core tasks, and application entities. AD datasets primarily consist of first-person perspective videos captured by in-vehicle cameras. These datasets support algorithms for perception, prediction, planning, and decision-making to assist safe and intelligent driving. In contrast, ITS focuses on third-person perspective traffic surveillance videos or images, enabling traffic management authorities to monitor road conditions, detect abnormal events, and enhance road safety.

\subsubsection{Models and Datasets}
Before the advent of LMMs, various small models and datasets ~\cite{naphade20192019,snyder2019data,krajewski2018highd,yu2020bdd100k} were developed to advance AD technology.
With the emergence of LMMs, several studies ~\cite{tom2023reading,jain2024semantic,deruyttere2022talk2car,zhang2023study,qian2024nuscenes,malla2023drama,wu2023language,sima2023drivelm,marcu2023lingoqa,cao2024maplm} develop AD-specific multimodal datasets to train specialized models for enhanced intelligence, capability, generalization, and interactivity in autonomous driving applications.
Notably, the SUTD-Traffic dataset~\cite{xu2021sutd} contains complex traffic scene videos collected from online video platforms, volunteer-operated cameras, or on-board cameras, with 62,535 video question-answer pairs manually annotated. This dataset is designed to enhance the complex driving context reasoning capabilities of LMMs within the AD domain. Similarly, the WTS dataset~\cite{wts}, featuring over 1,200 pedestrian-centric videos with captions, enables fine-grained analysis of pedestrian and vehicle behaviors using multi-view recordings in AD applications. The MM-AU ~\cite{Fang_2024_CVPR} annotated 58 ego-view traffic accident video QA pairs for safe driving systems. The RoadSocial ~\cite{Parikh_2025_CVPR} collected traffic videos from social media and constructed 260K QA pairs based on social media comments to enhance LMMs' understanding of road events.

These datasets differ significantly from our MITS data, making them unsuitable for directly improving LMM performance in ITS: (1) Domain difference. These datasets mainly aim to enhance LMM performance in autonomous driving, which is fundamentally different from ITS focusing on road monitoring and management. (2) Perspective difference. The data for autonomous driving primarily covers first-person perspectives like handheld and vehicle-mounted views, while MITS focuses on third-person surveillance camera perspectives. (3) Source difference. The data in related work mainly comes from open-source or internet data, whereas MITS data originates from non-public road surveillance cameras. (4) Task focus difference. As shown in \Cref{tab:relatedwork}, each dataset focuses on different task types (TT). TAD\cite{xu2022tad} and MM-AU\cite{Fang_2024_CVPR} focus on accident detection and accident reasoning. RAOD\cite{xu2024raod} specializes in road abandoned object detection. LingoQA\cite{marcu2023lingoqa} focuses on driving reasoning, object recognition, action justification, and scene description. SUTD\cite{xu2021sutd} focuses on 6 tasks: basic understanding, attribution, introspection, counterfactual inference, event forecasting, and reverse reasoning. WTS\cite{wts} focuses on location, attention, behavior, and context. RoadSocial\cite{Parikh_2025_CVPR} focuses on 11 tasks: description, consequence, key entity, where, why, temporal grounding, viewpoint, introspection, advisory, counterfactual, and adversarial. In contrast, the five key tasks in MITS are specifically designed to address the practical needs of traffic management authorities, emphasizing a holistic and supervisory understanding of traffic scenes.

\subsection{LMMs and Datasets for other Vertical Domains}
General-purpose large multimodal models demonstrate strong generalization and zero-shot task performance by leveraging self-supervised pretraining on large-scale image-text datasets, followed by supervised fine-tuning (SFT) with multimodal instruction-following data to align with human intents. Although these general LMMs ~\cite{zhu2023minigpt,liu2024visual,Qwen2VL,llava15-7b,qwen25-vl-7b,llava16-7b}, perform well across various tasks, their effectiveness tends to degrade in specialized areas where domain-specific data significantly differ from general web content. 
To overcome this, many efforts focus on developing specific LMMs tailored to specific domains. For example, Li et al.~\cite{li2024llava} and Bai et al.~\cite{bai2024m3d} develop specialized VQA datasets for the medical domain, enabling the training of medical-specific LMMs. Similarly, Liu et al.~\cite{liu2025multimodal} and Wei et al.~\cite{wei2024benchmarking} construct multimodal datasets and fine-tune general LMMs for tasks such as crop disease diagnosis. 

A key challenge in developing ITS-specific LMMs is the lack of high-quality multimodal data tailored to this domain.
To address this gap, we introduce MITS, the first large-scale multimodal benchmark dataset specifically designed for ITS. MITS includes 170,400 images across eight categories and 24 subcategories of objects and events, along with 5 million high-quality instruction-following image VQA pairs that span five core ITS tasks. Furthermore, we fine-tune mainstream general LMMs on MITS, significantly improving their performance and domain adaptation for ITS applications.

\section{MITS Benchmark}
\label{sec:MITS}
This section presents a comprehensive overview of the proposed MITS benchmark.
\Cref{sec:3.1} introduces the source images collected from real-world surveillance cameras and their statistics. \Cref{sec:3.2} describes the MITS dataset construction pipeline. \Cref{sec:3.3} presents statistical analyses of the constructed MITS dataset. Finally, \Cref{sec:3.4} presents the evaluation metrics employed in the MITS benchmark. 

\subsection{Source Images}
\label{sec:3.1}
\begin{figure}[!htbp]
  \centering
   \includegraphics[width=0.9\linewidth]{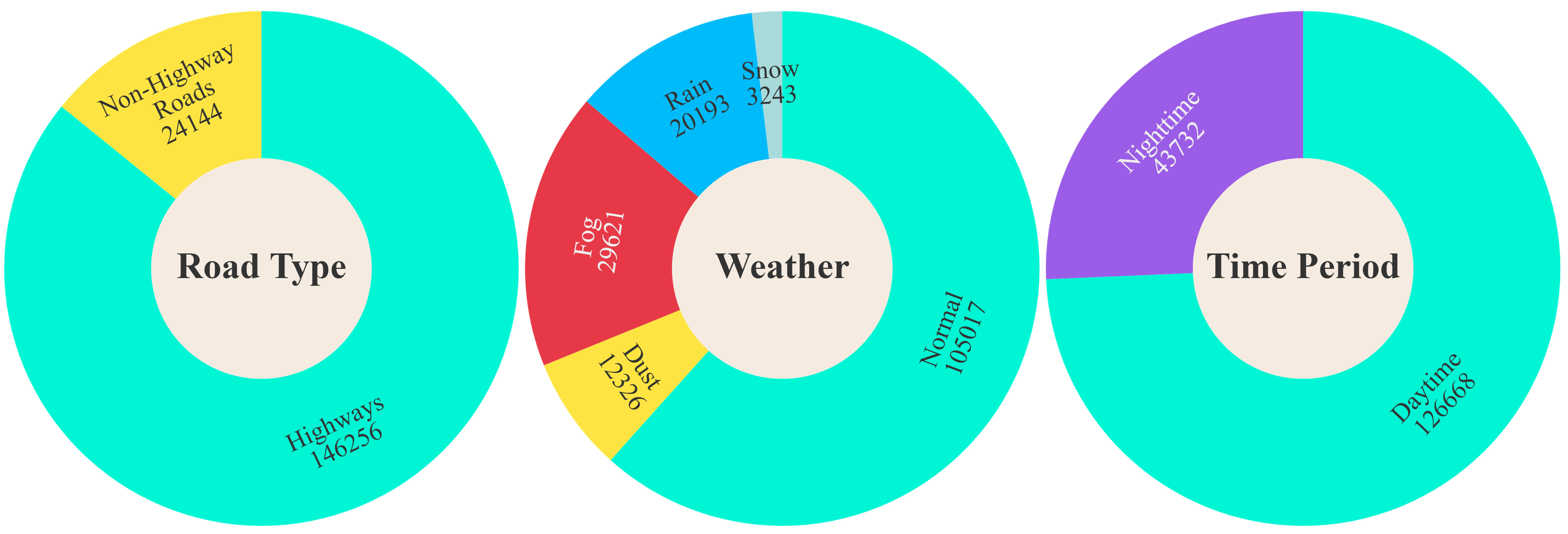}
   \caption{Base Statistics of the source images. The charts illustrate the distribution of images across road types, weather conditions, and time periods. Consistent with real-world ITS  scenarios, we intentionally avoid artificial balancing of environmental factors (road types, weather, time period) to maintain the natural occurrence frequencies observed in actual ITS scenarios.
}
   \label{fig:basestatistics}
\end{figure}

Over four years, we accumulate 170,400 high-quality, desensitized proprietary images from approximately 1,100 ITS surveillance cameras. Among these cameras, 80 are situated at intersections. 
These images are sourced from approximately 8000 video recordings, and most images have a resolution of 1920x1080. To build a representative image set for ITS applications, we manually sample frames that contained significant ITS-relevant events or objects, avoiding duplicates and ensuring comprehensive coverage of different scenarios. The dataset covers diverse road types (highways, and non-highway roads), weather conditions (dust, fog, rain, snow, normal), and time periods (daytime and nighttime), as shown in \Cref{fig:basestatistics}. 

\textbf{Privacy protection:} 
Ensuring data privacy and compliance is a paramount concern throughout the entire data lifecycle, from collection to release. We implement a multi-faceted strategy to ensure that no individual can be traced using our dataset:

\textbf{(1) Technical Desensitization of Personally Identifiable Information (PII).} We develop and apply a rigorous, automated desensitization pipeline to remove all potential PII from the images before their inclusion in the dataset. This pipeline specifically targets and anonymizes:

\textit{\textbf{License Plates}}: We choose a license plate detection model \cite{HyperLPR3} to identify all vehicle license plates. Once located, these areas are rendered unreadable through blur and pixelation.

\textit{\textbf{Human Faces}}: Similarly, any visible faces of pedestrians, drivers, or passengers are automatically detected using InsightFace \cite{Deng2020CVPR,insightface} and anonymized using the same robust blurring and pixelation techniques to prevent identification.

\textit{\textbf{Location-Specific Text Overlays}}: Many surveillance images include text overlays indicating camera IDs, intersection names, or other location-specific identifiers. Our pipeline also employs text detection algorithms \cite{easyocr} to identify these regions. The identified text is then blurred to prevent the disclosure of the cameras' precise geographical locations.

\textbf{(2) Procedural and Regulatory Compliance.} The entire process is conducted in strict adherence to relevant data protection laws and regulations. All data is handled within a secure, isolated environment with strict access controls, ensuring that only authorized personnel directly involved in the anonymization process could access the raw data. The publicly released dataset exclusively contains the fully desensitized images.

\begin{figure*}[!htbp]
  \centering
   \includegraphics[width=1.0\textwidth]{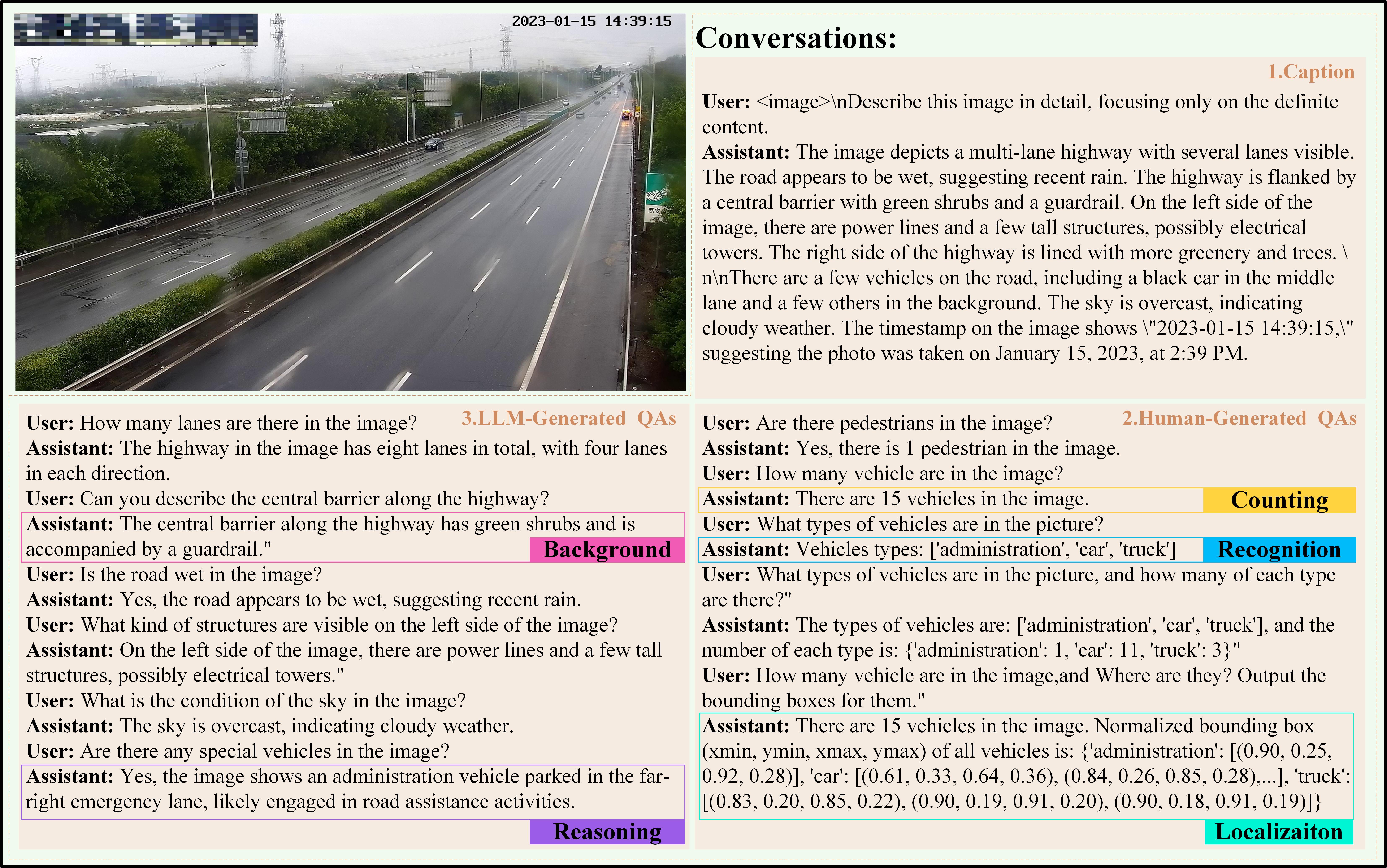}
   \caption{An example from MITS dataset, consisting of an image and three groups of QA pairs: Caption, Human-Generated QAs, and LLM-Generated QAs. These QA pairs cover five key ITS tasks: object and event \textbf{Recognition}, object \textbf{Counting}, object \textbf{Localization}, \textbf{Background} analysis, and event \textbf{Reasoning}}
   \label{fig:vqaexample}
\end{figure*}

\subsection{MITS Dataset Construction}
\label{sec:3.2}
An example of the MITS dataset is illustrated in \Cref{fig:vqaexample}, comprising of an image, Human-Generated QAs, LLM-Generated QAs, and Optimized caption. The construction of our MITS dataset follows three key steps: 

\textbf{(1) Objects \& Events Annotations:} First, we manually annotate the source images based on eight major categories and 24 subcategories most relevant to ITS scenarios. These annotations provide the foundational detection or classification labels for subsequent multimodal data generation.

\textbf{(2) Multimodal Tasks Definition:} Next, we define five ITS-specific multimodal tasks, designed to align with LMM applications in ITS scenarios.

\textbf{(3) Multimodal Data Generation:} Finally, we generate a set of VQA pairs for each image with a human-in-the-loop review process.

\subsubsection{Object \& Event Annotations}
\label{sec:3.2.1}
\textbf{Annotation categories:} 
Our \textbf{eight major categories} and \textbf{24 subcategories} stem from multiple years of ITS project experience, covering the most critical objects and events encountered in real applications. Specifically, these categories include anomaly conditions such as road abandoned object (spill), accident, fire and smoke (firesmoke), road construction, unusual weather, congestion (jam), and fundamental but indispensable objects like person and vehicle. This taxonomy ensures ITS-oriented coverage and is fully based on practical requirements for traffic surveillance, safety regulation, and incident management. The set of categories is defined as $S = \left\{ {{C_1},{C_2},...,{C_8}} \right\}$.

\textbf{Annotation Method:} A hybrid approach combining manual annotation with machine assistance is employed to generate annotations for small models. Initially, a subset of images is manually annotated, and this annotated set is used to train a YOLOv8~\cite{glenn_jocher_2022_7347926} model. The trained model then generates classification or detection labels for a new set of unannotated images. These machine-generated labels are subsequently reviewed and validated by human annotators from our expert team. The newly validated annotations are then incorporated into the training dataset to further refine and update the YOLOv8 model. This iterative process is repeated to ensure the accuracy and consistency of the annotations across all images.

\subsubsection{Multimodal Tasks Definition}
\label{sec:3.2.2}
Focusing on the core information extraction and practical problem-solving for ITS, we define five multimodal tasks:

\textit{\textbf{Object and Event Recognition}}: Questions that identify and recognize specific objects, scene elements, or events within an image. This task involves all object and event categories in $S$.

\textit{\textbf{Object Counting}}: Questions aimed at determining the number of specific objects in an image. This task involves object or event categories including road abandoned object, road construction, person, and vehicle.

\textit{\textbf{Object Localization}}: Questions requiring identification and description of object positions within an image, potentially involving coordinates or relative positioning. This task involves object or event categories including road abandoned object, accident, fire and smoke, road construction, person, and vehicle.

\textit{\textbf{Background Analysis}}: Questions focused on background elements in the image, assessing the model’s understanding of contextual and environmental information. This task involves unusual weather, traffic congestion, and other information elements not involved in $S$.

\textit{\textbf{Event Reasoning}}: Questions that demand logical analysis based on both background and foreground content in an image, involving causal relationships, multi-event correlation, intent analysis and potential event prediction. 

Each QA pair is categorized into one of the predefined tasks, as illustrated in \Cref{fig:vqaexample}. While these task names may appear generic, each QA instance is specifically designed to address ITS-specific conditions and semantics. For instance, the QA pair ("What's the cause of the traffic congestion?"-"A truck overturned, causing the traffic congestion.") falls under the event reasoning category, enabling supervisors to quickly identify abnormal situations and their root causes, thereby facilitating practical ITS problem-solving. This distinguishes it from other reasoning tasks, such as those in SUTD-TrafficQA ~\cite{xu2021sutd} and MM-AU ~\cite{Fang_2024_CVPR}, which primarily focus on autonomous driving decisions. From basic recognition to advanced reasoning, our hierarchical task pipeline systematically addresses challenges in real-world ITS deployments.

\subsubsection{Multimodal Data Generation}
\label{sec:3.2.3}
\begin{figure*}[!htbp]
  \centering
   \includegraphics[width=1.0\linewidth]{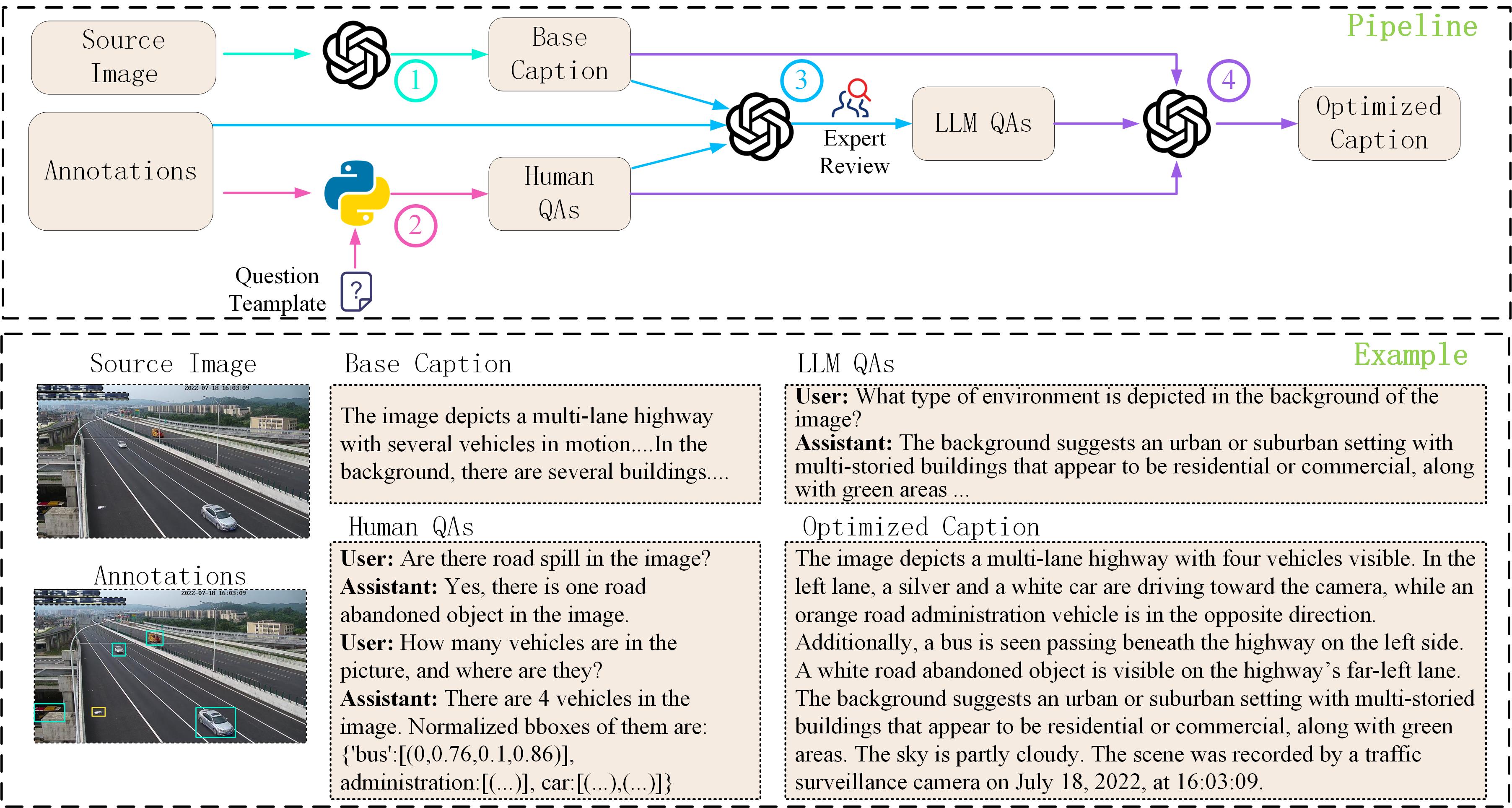}
   \caption{The MITS data generation pipeline.}
   \label{fig:generation}
\end{figure*}

Based on the source images and annotations, we design a human-in-the-loop approach to build high-quality MITS dataset, as shown in \Cref{fig:generation}.

\textit{1. Base Caption Generation:} We use GPT-4o ~\cite{gpt4o} to generate concise image captions. The prompt instructs the model to only include deterministic information from the images. To ensure the quality of these generated captions, an initial expert review is conducted by sampling 1000 images from all subcategories and assessing the resulting captions to establish an effective prompt. 
Captions generated through this process primarily include background details while containing limited foreground information.

\textit{2. Human-Generated QA Pairs Generation:} 
The process of generating human-generated QA pairs involves the following four steps:

(1) The first step is to create a question set for each object category. For each category ${C_i}$, we manually design a set of question templates covering three tasks: Recognition, Counting, and Localization. These templates incorporate various query formats, including Yes/No, What, Where, How many, and Which, ensuring diverse and comprehensive question formulations. The generated question set ${Q_i}$ focuses on foreground information specific to ${C_i}$ and are carefully crafted to ensure that accurate answers can be automatically derived from the manual annotations in the source data. The complete set of human-generated questions is represented as: $Q = \left\{ {{Q_1},{Q_2},...,{Q_i},...,{Q_8}} \right\}$, where ${Q_i} = \left\{ {q_i^1,q_i^2,...,q_i^{{N_i}}} \right\}$ and ${N_i}$ is the number of questions for ${C_i}$. 

(2) For each image ${p_j}$, we determine whether it is a positive or negative sample for each category ${C_i}$. An image ${p_j}$ is considered a positive sample for ${C_i}$ if it includes annotations related to ${C_i}$ in the source data; Otherwise, it is a negative sample. Then, we denote the set of positive images for category ${C_i}$ as $P{S_i}$, and the set of negative images for ${C_i}$ is  $N{S_i}$.

(3) The third step is to construct the image sample set ${S_i}$ for each category ${C_i}$. For each category ${C_i}$, we first create the positive sample set $S_i^P$ including all positive images, i.e., $S_i^P = P{S_i}$. Then, to ensure dataset balance, we randomly select an equal number of images from $N{S_i}$ to construct the negative images set $S_i^N \subseteq N{S_i}$, where $\left| {S_i^N} \right| = \left| {S_i^P} \right|$. If $\left| {N{S_i}} \right| < \left| {P{S_i}} \right|$, then $S_i^N$ is set to $N{S_i}$. Note that an image ${p_j}$ can be a positive sample for several categories simultaneously.

(4) The last step is to construct the human questions and answers set $Q{A_{{p_j}}}$ for each image ${p_j}$. If ${p_j} \in S_i^P$ or ${p_j} \in S_i^N$ for a given ${C_i}$, we retrieve all questions from ${Q_i}$ and generate accurate answers based on the source data annotations using python scripts:
\begin{equation}
  Q{A_{{p_j}}} = \bigcup\limits_{{C_i};{p_j} \in S_i^P \cup S_i^N} {\left\{ {(q,a)\left| {q \in {Q_i}} \right.} \right\}}
  \label{eq:important}
\end{equation}


\textit{3. LLM-Generated QA Pairs Generation:} 
While human-generated QAs effectively cover tasks such as Recognition, Counting, and Localization by leveraging reliable foreground annotations, they often provide insufficient coverage of background-related elements. To address this gap, we employ GPT-4o to automatically generate QA pairs specifically targeting Background Analysis and Event Reasoning.

Inspired by LLaVA, we craft detailed prompt for GPT-4o using base caption, human-generated QA pairs, and annotation data, instructing the model to produce deterministic (i.e., clearly positive or negative) QA pairs. The prompt undergo iterative refinement based on an initial expert review, minimizing hallucinations and ensuring higher reliability. Finally, all LLM-generated QA pairs pass through an expert review process to remove factual errors or ambiguous reasoning, thereby maintaining the overall quality of our dataset. The prompt is: 

\begin{promptbox}{The Prompt for Generating LLM QAs}
        You are an AI visual assistant, and you are seeing a single image with a caption, many human-written QA pairs and human-annotated object/event labels.
        The input caption describes the image you are looking at, and the input human-written QA pairs and human-annotated labels describes the important foreground objects and events on the image.
        The input human-written QA pairs and human-annotated labels are reliable, all of them are related to the recognition and location for foreground objects and events in Intelligent Traffic Suverilliance Scenario.
        
        \textbf{Your Tasks:}
        
        \textbf{Task1:} Write some new QA pairs related to "Background Analysis" following the below principles:
        (1) Only write QA pairs focusing on the background information - not related to the foreground objects and events in the input human-written QAs.
        (2) Only include questions that have definite answers, definitely positive or definitely negative.
        (3) The writing style should be diverse, Don't just generate Yes or No format QA.
        (4) You must write them in English.
        \textbf{Task2:} Write some new QA pairs related to "Event Reasoning" following the below principles:
        (1) Only write QA pairs focusing on the Event Reasoning. Event inference refers to the process of using known information, data, or premises to derive information not directly contained in the input through logical analysis, empirical analysis, statistical analysis, etc. This includes analyses such as causal analysis, relational analysis, relative position analysis, intent analysis, and more.
        (2) You should make the best of the input caption and Human-Written QAs.
        (3) The writing style should be diverse, Don't just generate Yes or No format QA.
        (4) You should generate reliable reasoning.
        (5) You should generate some difficult question and reliable answer.
        (6) You must write them in English.

        \textbf{Your Output Format:}
        Finally, you need to output a json data containing all QA pairs in task1 and task2, and the output format is JSON Format. Please only output JSON Content:
        [
            \{
                Question: ***,
                Answer: ***,
                Task Type: ***
            \},
            \{
                Question: ***,
                Answer: ***,
                Task Type: ***
            \},
            .....
        ]
        
        In your output, the Task Type for task1 is "Background Analysis", and for the QA in task2, its Task Type is "Event Reasoning".

\end{promptbox}

\textit{4. Optimized Caption:} 
The base captions typically contain unverified background information (lacking expert review) and provide limited foreground details. Conversely, human-generated QAs offer precise foreground information, and LLM-generated QAs contain expert-reviewed, accurate background content. By integrating both human- and LLM-generated QAs into the base captions, we create high-quality captions enriched with comprehensive foreground and background details. To achieve this, we employ a carefully designed prompt to guide GPT-4o in synthesizing existing information into refined, optimized captions. The prompt is:

\begin{promptbox}{The Prompt for Generating Optimized Caption}

\textbf{Task:} 

Generate an optimized image caption by strategically integrating information from multiple sources.

\textbf{Input Information Sources:}

1. Base Caption: The initial caption generated by GPT-4o, potentially containing unverified or incomplete information.

2. Human-Generated QAs: Expert-created QAs providing reliable foreground object and event descriptions.

3. LLM-Generated QAs: LLM-generated supplementary background QAs that has undergone human verification.

\textbf{Processing Rules:}

1. Apply strict reliability hierarchy when resolving information conflicts: Human QAs > LLM QAs > Base Caption.

2. Include all key information in human QAs and LLM QAs.

3. Maintain natural language fluency while maximizing information density.

\textbf{Output Format:}

{

    "Optimized Caption": "*****"
    
}
\end{promptbox}

\subsection{MITS Dataset Statistics}
\label{sec:3.3}
\subsubsection{Image Statistics}
\label{sec:3.3.1}
\begin{figure}[!htbp]
  \centering
   \includegraphics[width=1.0\linewidth]{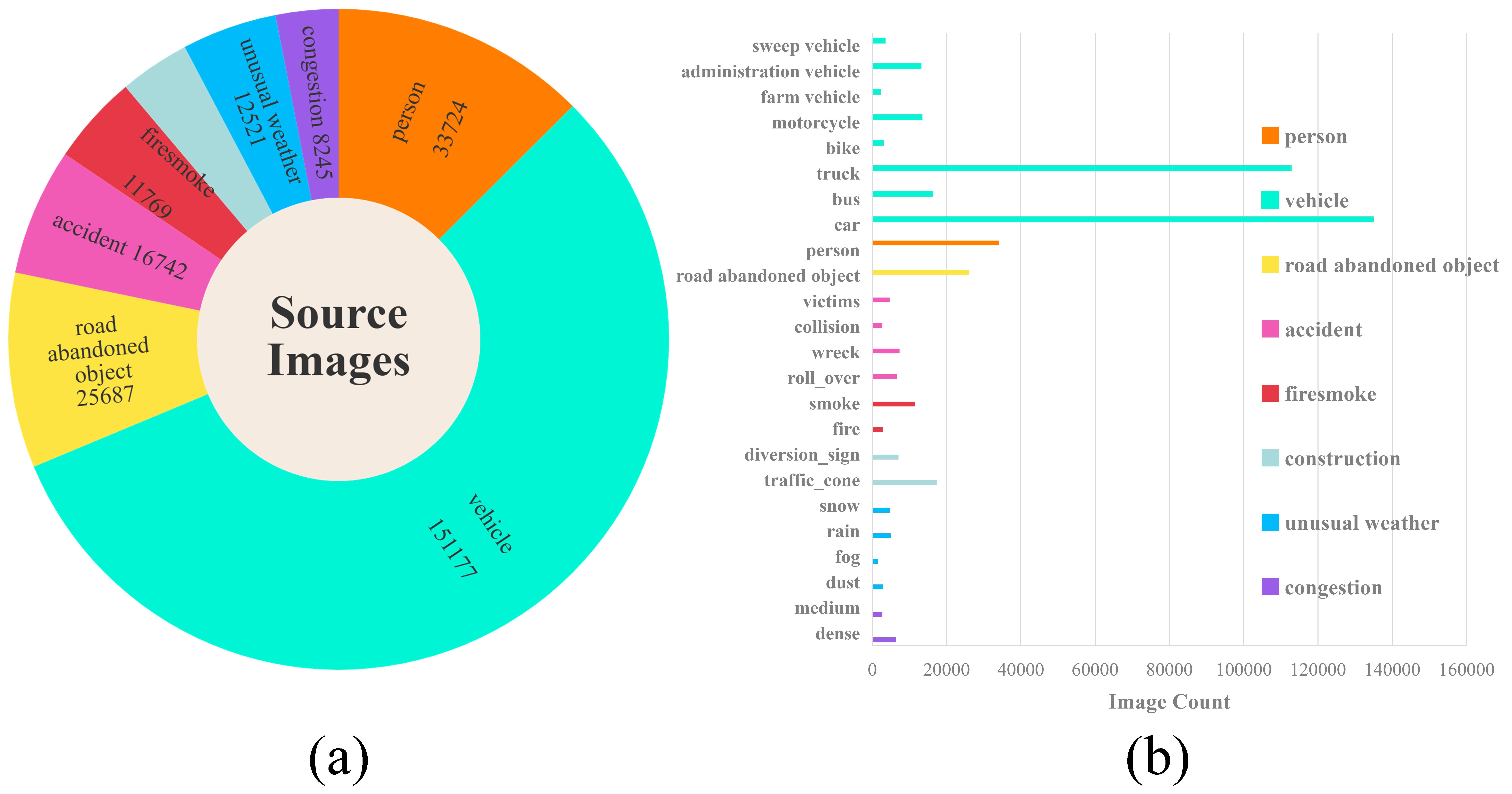}
   \caption{(a) Distribution and quantity of images across the 8 main categories. (b) Number of images containing objects from the 24 subcategories. The highest frequency of objects pertains to cars and trucks, reflecting their prominence in ITS, particularly on highways. Other subcategories are represented with a relatively balanced distribution.}
   \label{fig:categorywise_statistics}
\end{figure}

\Cref{fig:categorywise_statistics} shows the distribution of images containing each category and subcategory. It is important to note that a single image may feature objects or events from multiple categories. \Cref{tab:categorywise_basestatistics} breaks down image distributions for each category by road type, weather condition, and time period. To ensure diversity, we uniformly sampled 10\% of images for the test set, balancing object subcategories, road types, weather conditions, and time periods.

\begin{table*}[htbp]
  \centering
  \resizebox{\textwidth}{!}{
    \begin{tabular}{cc|cc|ccccc|cc}
    \toprule
          &       & \multicolumn{2}{c|}{\textbf{Road Type}} & \multicolumn{5}{c|}{\textbf{Weather Condition}} & \multicolumn{2}{c}{\textbf{Time Period}} \\
    \textbf{Category} & \textbf{Sum} & \textbf{Highways} & \textbf{Others} & \textbf{Dust} & \textbf{Fog} & \textbf{Rain} & \textbf{Snow} & \textbf{Normal} & \textbf{Day} & \textbf{Night} \\
    \midrule
    \textbf{person} & 33724 & 27307 & 6417  & 2337  & 3970  & 2584  & 260   & 24573 & 26500 & 7224 \\
    \textbf{vehicle} & 151177 & 132543 & 18634 & 9610  & 25987 & 18697 & 2088  & 94795 & 116742 & 34435 \\
    \textbf{abandoned object} & 25687 & 25055 & 632   & 412   & 1947  & 1447  & 5     & 21876 & 20087 & 5600 \\
    \textbf{accident} & 16742 & 7725  & 9017  & 2450  & 1997  & 3264  & 16    & 9015  & 14346 & 2396 \\
    \textbf{firesmoke} & 11769 & 10173 & 1596  & 3460  & 2664  & 196   & 20    & 5429  & 10763 & 1006 \\
    \textbf{construction} & 9267  & 9236  & 31    & 142   & 876   & 276   & 2     & 7971  & 4824  & 4443 \\
    \textbf{unusual weather} & 12521 & 9471  & 3050  & 2231  & 3789  & 3681  & 2704  & 116   & 7931  & 4590 \\
    \textbf{congestion} & 8245  & 7961  & 284   & 125   & 1252  & 1265  & 18    & 5585  & 6664  & 1581 \\
    \bottomrule
    \end{tabular}%
    }
    \caption{Distribution of images by category across road types, weather conditions, and time periods.}
  \label{tab:categorywise_basestatistics}%
\end{table*}%


\subsubsection{VQA Statistics}
\label{sec:3.3.2}
\begin{figure}[!htbp]
  \centering
   \includegraphics[width=1.0\linewidth]{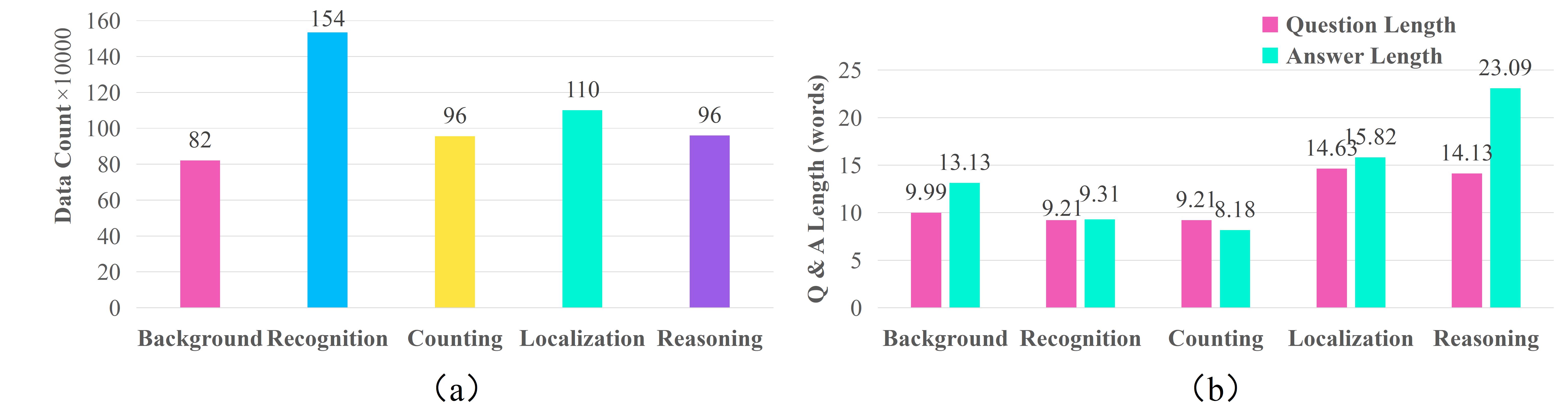}
   \caption{Statistics of QA pairs in MITS. (a) Number of QAs for each task. (b) Average question and answer length per task.}
   \label{fig:vqastatistics}
\end{figure}

Using the generation pipeline in \Cref{sec:3.2}, we generate more than 5,373,391 VQA pairs, covering the defined five multimodal tasks. \Cref{fig:vqastatistics} shows the distribution of QA pairs per task and the average lengths of questions and answers.

\subsection{Evaluation Metrics} 
\label{sec:3.4}
To comprehensively evaluate model performance across the diverse ITS tasks, we employ a set of task-specific metrics. The selection of these metrics is guided by industry standards and the unique characteristics of each task's expected output.

\textbf{\textit{$S_{loc}$:}} For the Localization task, we use the Intersection over Union (IoU) between the predicted bounding box ($B_{pred}$) and ground truth bounding box ($B_{gt}$) as the localization score $S_{loc}$, as shown in \cref{eq:iou}. Its value ranges from 0 (no overlap) to 1 (perfect alignment), providing a robust and intuitive measure of localization accuracy.

\begin{equation}
S_{loc} = {{Area\left( {{B_{pred}} \cap {B_{gt}}} \right)} \over {Area\left( {{B_{pred}} \cup {B_{gt}}} \right)}}
  \label{eq:iou}
\end{equation}

\textbf{\textit{$S_{count}$:}} For the Counting task, we compute the counting score $S_{count}$ based on the Relative Absolute Error (RAE), as shown in \cref{eq:rae}. We chose RAE over absolute error because object counts in our dataset can vary significantly. This ensures that an error of the same magnitude is penalized less for images with high object counts, leading to a more balanced and fair assessment across all samples.

\begin{equation}
{S_{count}} = \max (0,1 - {{\left| {{N_{pred}} - {N_{gt}}} \right|} \over {{N_{gt}}}})
  \label{eq:rae}
\end{equation}

\textbf{\textit{$S_{recog}$:}} For the Recognition task, which often involves "Yes/No" or categorical identification questions, we use the accuracy as the recognition score $S_{recog}$. This metric is the most direct and suitable measure, calculating the proportion of predictions that exactly match the ground-truth answers.

\textit{$S_{background}$ and $S_{reasoning}$:} The answers to Background analysis and Reasoning tasks are generative, free-form text, making exact-match metrics like accuracy inadequate. So we follow the methodology of \cite{lian2024best,liu2024visual}, utilizing DeepSeek-R1 \cite{deepseekr1} as a scoring model to evaluate alignment between model outputs and ground truth answers, with scores normalized from 0 to 1. The scoring prompt is: 

\begin{promptbox}{DeepSeek-R1 Scoring Prompt}
You are tasked with evaluating and scoring predictions against a standard answer. The input will consist of a set of (Question, Standard Answer, Question Type, Prediction). Your job is to evaluate and score the prediction according to how closely it matches the standard answer, using the following criteria:

1. Conduct factual alignment check between the Prediction and Standard Answer, verifying all critical entities.

2. Evaluate the semantic similarity between the prediction and the standard answer.

3. Analyze the contextual appropriateness, logical coherence, and topical relevance demonstrated in the Prediction.

Scores should be given on a scale from 0 to 1, where 1.00 indicates perfect accuracy and 0.00 indicates no match. Output format is:

{

    "Score for Prediction": score
    
}
\end{promptbox}

Finally, we perform human verification on random sample of 5,000 QA pairs. Three experts independently analyze the thinking process of DeepSeek-R1 and assess whether the automated scores are reasonable. 98\% of the automatic scoring results are regarded as reasonable by all the experts.

\section{Experiments}
\label{sec:experiments}
In this section, we present comprehensive experiments on the MITS dataset to validate its effectiveness and value in enhancing model recognition and understanding capabilities for ITS scenarios.
\subsection{Experiment Setup}
\subsubsection{Experiment Content} 
To date, there are no LMMs specifically designed for ITS, so we select mainstream general-domain LMMs to conduct our experiments. This approach simultaneously demonstrates MITS's broader value for general LMM communities.

\textbf{(1) General LMM Evaluation:} We evaluate the performance of original general LMMs on the MITS test dataset as a baseline.

\textbf{(2) Model Fine-Tuning:} Following the training methodology outlined in LLaVA-Med~\cite{li2024llava} and CDDM~\cite{liu2025multimodal}, we fine-tune general LMMs using multi-round conversations on the training set. This process includes ITS-specific image-text alignment training using captions and instruction tuning with SFT on all QA pairs. To optimize training efficiency, we froze the vision encoder and fine-tuned only the projector and the language model using LoRA. The fine-tuned models are subsequently evaluated on the test set.

\textbf{(3) Comparative Studies:} We perform extensive comparative experiments to validate the dataset's effectiveness and applicability.

\subsubsection{Implementation Details} 
We select 7 widely used LMMs for our experiments, and all experiments are conducted on 16x NVIDIA H100 (80GB) GPUs. 
We utilize Pytorch framework along with the Hugging Face Transformers library and MS-Swift Toolkit for model implementation and training. Implementation details and key hyperparameters are detailed in \Cref{tab:hypara}.

\begin{table}[!ht]
    \centering
    \caption{Implementation details and key hyperparameters. We train five distinct models and evaluate seven models across different architectures and scales. For training, we maintain consistent hyperparameters including batch size (BS), learning rate (LR), learning rate scheduler (LRS), weight decay (WD), and training epochs. During evaluation, we set the maximum new sequence length (MSL) to 1024 tokens and maximum image resolution (MR) to 1920×1080 pixels.}
    \resizebox{\textwidth}{!}{
    \begin{tabular}{cccccccccc}
    \hline
        \textbf{Model} & \textbf{Evaluate} & \textbf{Train} & \textbf{BS} & \textbf{LR} & \textbf{LRS} & \textbf{WD} & \textbf{Epoch} & \textbf{MSL} & \textbf{MR} \\ \hline
        LLaVA1.5-7B \cite{llava15-7b} & \Checkmark & \Checkmark & 32 & 2e-4 & cosine & 0.1 & 2 & 1024 & 1920x1080 \\ 
        LLaVA1.6-7B \cite{llava16-7b} & \Checkmark & \Checkmark & 32 & 2e-4 & cosine & 0.1 & 2 & 1024 & 1920x1080 \\ 
        Qwen2-VL-7B \cite{qwen2-vl-7b} & \Checkmark & \Checkmark & 16 & 1e-4 & cosine & 0.1 & 2 & 1024 & 1920x1080 \\ 
        Qwen2.5-VL-7B \cite{qwen25-vl-7b} & \Checkmark & \Checkmark & 16 & 1e-4 & cosine & 0.1 & 2 & 1024 & 1920x1080 \\ 
        Qwen2.5-VL-3B \cite{qwen25-vl-3b} & \Checkmark & \Checkmark & 16 & 1e-4 & cosine & 0.1 & 2 & 1024 & 1920x1080 \\ 
        Qwen2.5-VL-32B \cite{qwen25-vl-32b} & \Checkmark & / & / & / & / & / & / & 1024 & 1920x1080 \\ 
        Yi-VL-34B \cite{Yi-VL-34B} & \Checkmark & / & / & / & / & / & / & 1024 & 1920x1080 \\ \hline
    \end{tabular}
    }
    \label{tab:hypara}
\end{table}

In addition to the training hyperparameters, we clarify our handling of localization data to ensure valid training and evaluation. Bounding box coordinates in the MITS dataset are provided in a normalized [0, 1] floating-point range, following the (x\_min, y\_min, x\_max, y\_max) format, which is derived by dividing raw pixel values by the image's corresponding dimensions.

To accommodate different model architectures, our fine-tuning pipeline employs a model-specific preprocessing step. For models that require absolute coordinates, such as Qwen2-VL and Qwen2.5-VL, we convert the normalized values back to absolute pixel coordinates before training. Conversely, for models like LLaVA-1.5 and LLaVA-1.6 that natively process normalized data, the coordinates are used directly. To maintain consistency during evaluation, all predicted bounding boxes are unified to the [0, 1] range before IoU calculation. This model-aware process guarantees that localization data is correctly aligned for both training and evaluation across all architectures.

\begin{table*}[!htbp]
  \centering
    \caption{Scores on each task's test dataset before and after fine-tuning with different training dataset components. }
  \resizebox{\textwidth}{!}{
    \begin{tabular}{cccccccc}
    \toprule
    \textbf{Method} & \multicolumn{1}{c}{\textbf{SFT Data}} & \textbf{Background} & \textbf{Recognition} & \textbf{Counting} & \textbf{Localization} & \textbf{Reasoning} & \textbf{Average} \\
    \midrule
    LLaVA-1.5-7B\cite{llava15-7b} & /     & 0.612 & 0.589 & 0.483 & 0.078 & 0.737 & 0.494 \\
    LLaVA-1.5-7B & BC+QA & 0.802 & 0.926 & 0.824 & 0.881 & 0.816 & 0.860 \\
    LLaVA-1.5-7B & OC+QA & 0.858 & 0.958 & 0.877 & 0.908 & 0.884 & \textbf{0.905(+83.2\%)} \\
    \midrule
    LLaVA-1.6-7B\cite{llava16-7b} & /     & 0.701 & 0.755 & 0.520 & 0.620 & 0.760 & 0.678 \\
    LLaVA-1.6-7B & OC+QA & 0.877 & 0.967 & 0.904 & 0.936 & 0.886 & \textbf{0.921(+35.8\%)} \\
    \midrule
    Qwen2-VL-7B\cite{qwen2-vl-7b} & /     & 0.729 & 0.778 & 0.594 & 0.051 & 0.760 & 0.584 \\
    Qwen2-VL-7B & BC+QA & 0.843 & 0.938 & 0.866 & 0.892 & 0.836 & 0.883 \\
    Qwen2-VL-7B & OC+QA & 0.875 & 0.972 & 0.914 & 0.945 & 0.889 & \textbf{0.926(+58.6\%)} \\
    \midrule
    Qwen2.5-VL-3B\cite{qwen25-vl-3b} & /     & 0.730 & 0.735 & 0.534 & 0.113 & 0.782 & 0.578 \\
    Qwen2.5-VL-3B & OC+QA & 0.858 & 0.955 & 0.906 & 0.932 & 0.830 & \textbf{0.904(+56.4\%)} \\
    \midrule
    Qwen2.5-VL-7B\cite{qwen25-vl-7b} & /     & 0.787 & 0.798 & 0.665 & 0.553 & 0.854 & 0.732 \\
    Qwen2.5-VL-7B & OC+QA & 0.889 & 0.973 & 0.915 & 0.949 & 0.890 & \textbf{0.930(+27\%)} \\
    \midrule
    Qwen2.5-VL-32B\cite{qwen25-vl-32b} & /     & 0.840 & 0.786 & 0.656 & 0.583 & 0.882 & 0.746 \\
    \midrule
    Yi-VL-34B\cite{Yi-VL-34B} & /     & 0.606 & 0.648 & 0.385 & 0.044 & 0.677 & 0.475 \\
    \bottomrule
    \end{tabular}%
    }
  \label{tab:overall}%
\end{table*}%

\subsection{Experiment Results}
\subsubsection{Overall Performance} 
\Cref{tab:overall} summarizes the overall performance of models on the MITS test set before and after fine-tuning with different training dataset components.

In the ``SFT Data" column of \Cref{tab:overall}, ``/" represents the performance of the original models. ``BC+QA" indicates training with base captions and other QA pairs, while ``OC+QA" refers to training with optimized captions and other pairs. The results show:

1. The introduction of the MITS dataset results in significant improvements in model performance within the ITS domain, particularly for object counting and localization tasks, highlighting its ability to enhance models' alignment and understanding of ITS scenarios.

2. Models trained with optimized captions (OC+QA) outperform those trained with base captions (BC+QA), demonstrating that optimized captions offer more reliable and comprehensive information.

3. The primary performance bottleneck lies in data rather than model design or model scale. While LLaVA-1.6 shows a 37.2\% improvement over LLaVA-1.5, MITS-finetuned LLaVA-1.5 achieves an 83.2\% performance boost. The consistent pattern is observed in Qwen-7B models (58.6\% vs 25.3\% improvement).

\subsubsection{Category-wise Performance} 
\Cref{tab:category_performance} illustrates the category-wise evaluation results for original and fine-tuned models (trained with OC+QA). Across all categories, performance consistently improves after fine-tuning, though the improvement magnitude varies. As observed from the performance before and after fine-tuning, among all categories, unusual weather and traffic congestion are relatively simpler, while construction, person and accident categories are comparatively more challenging.

\begin{table}[!ht]
    \centering
    \caption{Category-wise performance comparison before and after fine-tuning on MITS. The fine-tuning process significantly improves performance across all eight major ITS categories.}
    \label{tab:category_performance}
      \resizebox{\textwidth}{!}{
    \begin{tabular}{ccccccccccc}
    \hline
        \textbf{Method} & \textbf{SFT} & \textbf{person} & \textbf{vehicle} & \textbf{spill} & \textbf{accident} & \textbf{firesmoke} & \textbf{construction} & \textbf{weather} & \textbf{jam} & \textbf{average} \\ \hline
        LLaVA-1.5-7B\cite{llava15-7b} & / & 0.464 & 0.491 & 0.492 & 0.490 & 0.450 & 0.499 & 0.534 & 0.516 & 0.489 \\ 
        LLaVA-1.5-7B & \Checkmark & 0.882 & 0.902 & 0.904 & 0.875 & 0.892 & 0.898 & 0.912 & 0.925 & 0.899 \\ 
        \midrule
        LLaVA-1.6-7B\cite{llava16-7b} & / & 0.583 & 0.670 & 0.646 & 0.626 & 0.613 & 0.614 & 0.739 & 0.736 & 0.653 \\ 
        LLaVA-1.6-7B & \Checkmark & 0.903 & 0.920 & 0.923 & 0.894 & 0.911 & 0.909 & 0.927 & 0.933 & 0.916 \\ 
        \midrule
        Qwen2-VL-7B\cite{qwen2-vl-7b} & / & 0.556 & 0.578 & 0.568 & 0.557 & 0.536 & 0.597 & 0.620 & 0.617 & 0.576 \\ 
        Qwen2-VL-7B & \Checkmark & 0.910 & 0.925 & 0.934 & 0.895 & 0.915 & 0.919 & 0.931 & 0.935 & 0.922 \\ 
        \midrule
        Qwen2.5-VL-3B\cite{qwen25-vl-3b} & / & 0.540 & 0.575 & 0.576 & 0.553 & 0.540 & 0.580 & 0.594 & 0.595 & 0.569 \\ 
        Qwen2.5-VL-3B & \Checkmark & 0.885 & 0.902 & 0.913 & 0.866 & 0.893 & 0.893 & 0.912 & 0.914 & 0.899 \\ 
        Qwen2.5-VL-7B\cite{qwen25-vl-7b} & / & 0.643 & 0.724 & 0.733 & 0.674 & 0.650 & 0.683 & 0.774 & 0.804 & 0.710 \\ 
        Qwen2.5-VL-7B & \Checkmark & 0.914 & 0.928 & 0.940 & 0.899 & 0.919 & 0.923 & 0.933 & 0.935 & 0.926 \\ 
        Qwen2.5-VL-32B\cite{qwen25-vl-32b} & / & 0.669 & 0.740 & 0.716 & 0.700 & 0.674 & 0.704 & 0.801 & 0.821 & 0.726 \\ 
        \midrule
        Yi-VL-34B\cite{Yi-VL-34B} & / & 0.443 & 0.474 & 0.455 & 0.479 & 0.438 & 0.498 & 0.481 & 0.518 & 0.470 \\ \hline
    \end{tabular}
    }
\end{table}

\subsubsection{Qualitative Analysis}
Taking the Qwen2.5-VL-7B model as an example, we visualize the performance differences before and after training on MITS, qualitatively analyze the performance improvements brought by our dataset and discuss failure cases along with underlying reasons.

\begin{figure}[!htbp]
  \centering
   \includegraphics[width=1\linewidth]{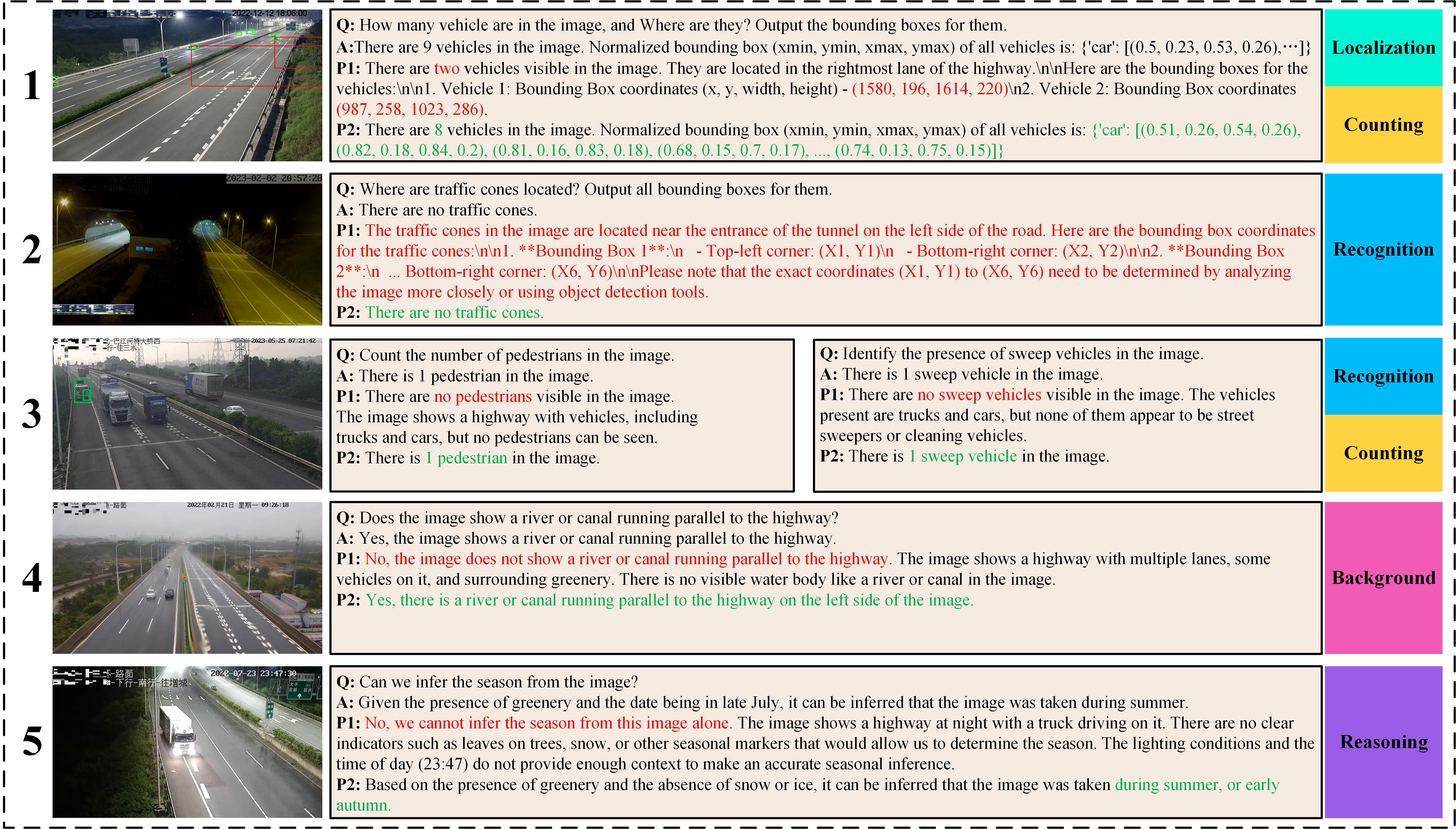}
   \caption{Representative success cases across five tasks. Text in red indicates incorrect responses from the baseline model, while green highlights correct answers generated by the fine-tuned model.}
   \label{fig:show_goodcase}
\end{figure}

By selecting test samples where the fine-tuned model significantly outperforms the original model, we curated five representative case studies (one for each task) and present them in \Cref{fig:show_goodcase}.

Case 1 (Object Localization \& Counting):
The baseline model failed dramatically in vehicle counting and localization, detecting only 2 vehicles (incorrectly) with bounding boxes extending beyond image boundaries (red rectangles). After fine-tuning, the model correctly identified 8 vehicles with accurate spatial localization (green boxes), demonstrating significant improvement despite missing one instance.

Case 2 (Recognition):
The baseline model exhibited hallucination by falsely detecting non-existent traffic cones. Fine-tuning effectively mitigated this hallucination problem, showing improved visual recognition capability.

Case 3 (Object Counting \& Recognition):
For pedestrian detection (left), the fine-tuned model corrected the baseline's failure to recognize the pedestrian in highway. For vehicle recognition (right), it acquired the ability to identify rare vehicle types (e.g., street sweeper) that were beyond the baseline's recognition capacity.

Case 4 (Background Analysis):
The baseline model failed to recognize a partially occluded river in the image periphery due to limited contextual understanding. The fine-tuned version demonstrated enhanced scene comprehension by successfully identifying this secondary background element.

Case 5 (Reasoning):
While not matching ground truth labels that used timestamp information, the fine-tuned model displayed emerging reasoning capability by correctly inferring the summer season from vegetation characteristics, unlike the baseline which lacked such analytical capacity.

\begin{figure}[!htbp]
  \centering
   \includegraphics[width=1\linewidth]{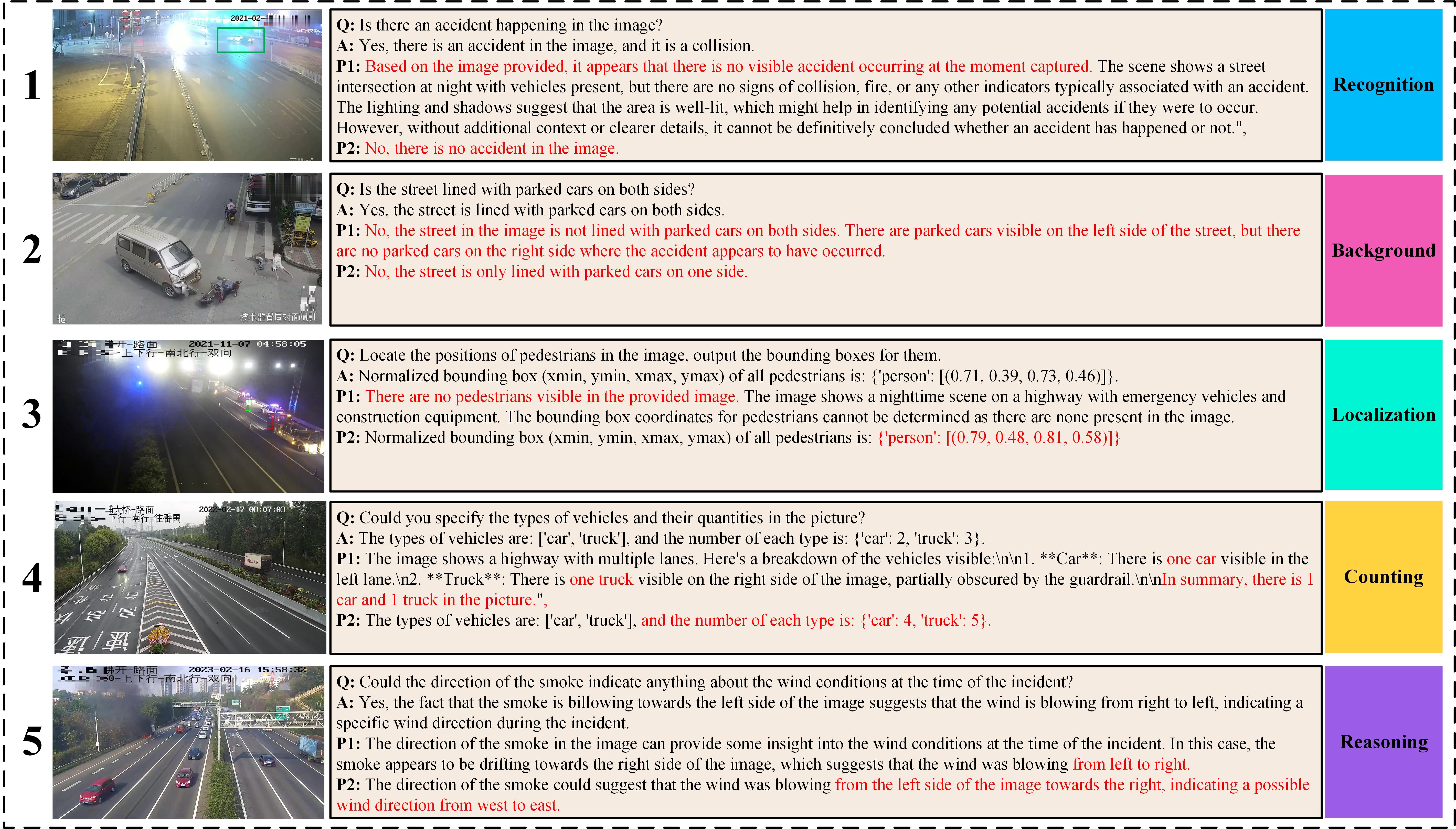}
   \caption{Representative failure cases across five tasks. Red text highlights erroneous responses that persist even after model fine-tuning, demonstrating remaining challenges in ITS scenario.}
   \label{fig:show_badcase}
\end{figure}

While MITS enhances the performance of LMMs in ITS scenarios, the fine-tuned models still struggle with certain challenging cases. By analyzing instances where both the original and fine-tuned models achieve low scores, we identify persistent limitations and present five representative failure cases in \Cref{fig:show_badcase}.

Case 1 (Recognition):
Both baseline and fine-tuned models failed to detect a vehicle collision accident (green bounding box) in the input image. This limitation likely stems from challenging imaging conditions including limited camera perspective, poor nighttime illumination, glare effects, and motion blur - factors that collectively degrade recognition performance.

Case 2 (Background Analysis):
The models exhibited partial scene understanding, detecting vehicles in parking spaces on only one side of the road despite bilateral presence. This selective attention may result from the models' tendency to prioritize salient features (e.g., a prominent traffic accident) at the expense of comprehensive scene analysis.

Case 3 (Object Localization):
While the baseline model completely missed a pedestrian (green box), the fine-tuned version produced an incorrect bounding box (red box). This localization failure under low-light conditions suggests that illumination robustness remains a significant challenge for current architectures.

Case 4 (Object Counting):
Discrepancies emerged in vehicle counting: the baseline undercounted (missing 2 cars and 3 trucks), while the fine-tuned model overcounted. This pattern indicates that training on datasets with numerous small, distant objects may induce false positives for similar visual patterns.

Case 5 (Event Reasoning):
Both model versions incorrectly inferred wind direction from smoke movement (right-to-left), revealing fundamental limitations in deriving physical dynamics from static visual cues.

Challenging scenarios (low-light conditions, small/dense objects, occlusions, reflections) continue to pose significant difficulties, suggesting that MITS dataset fine-tuning alone may be insufficient for robust ITS performance (accuracy $>$ 98\%) - architectural improvements may be necessary. Meanwhile, single-view or single-frame inputs inherently limit recognition and reasoning capabilities, indicating potential value in developing: Multi-view correlated datasets, temporal video datasets, and Spatiotemporal modeling approaches.

\section{Conclusion}
\label{sec:conclusion}
To address the performance degradation of general-domain LMMs in ITS applications, we introduce MITS, including 170,400 ITS-specific images with captions and over 5M VQA pairs. Extensive experiments demonstrate that MITS enables remarkable performance improvements (27.0\%-83.2\%) when fine-tuning state-of-the-art LMMs, establishing its dual value as both: (1) a transformative resource for ITS applications, and (2) a pioneering case study in vertical-domain LMM adaptation. In the future, we plan to evaluate additional LMMs on our benchmark, expand the dataset with multi-camera correlated data and video-based multimodal data, and further optimize LMM architectures to better address ITS-specific challenges.



\bibliographystyle{elsarticle-num} 
\bibliography{reference}

\end{document}